# LVLMs as inspectors: an agentic framework for category-level structural defect annotation

**Sheng Jiang[1,2], Yuanmin Ning[2], Bingxi Huang[2], Peiyin Chen[3], Zhaohui Chen[4*]**

1. *National Key Laboratory of Water Disaster Prevention, Hohai University, Nanjing 210098, China,* sheng.jiang@hhu.edu.cn

2. *College of Water Conservancy and Hydropower Engineering, Hohai University, Nanjing 210098, China,* 203958475@qq.com; 240802010002@hhu.edu.cn

3. *College of Artificial Intelligence and Automation, Hohai University, Changzhou 213200, China,* pychen@hhu.edu.cn

4. *School of Civil Engineering, The University of Sydney, NSW 2006, Australia*

*Corresponding author:*  zhaohui.chen@sydney.edu.au

**Abstract:**

Automated structural defect annotation is essential for ensuring infrastructure safety while minimizing the high costs and inefficiencies of manual labeling. A novel agentic annotation framework, Agent-based Defect Pattern Tagger (ADPT), is introduced that integrates Large Vision-Language Models (LVLMs) with a semantic pattern matching module and an iterative self-questioning refinement mechanism. By leveraging optimized domain-specific prompting and a recursive verification process, ADPT transforms raw visual data into high-quality, semantically labeled defect datasets without any manual supervision. Experimental results demonstrate that ADPT achieves up to 98% accuracy in distinguishing defective from non-defective images, and 85%-98% annotation accuracy across four defect categories under class-balanced settings, with 80%-92% accuracy on class-imbalanced datasets. The framework offers a scalable and cost-effective solution for high-fidelity dataset construction, providing strong support for downstream tasks such as transfer learning and domain adaptation in structural damage assessment.

## 1 Introduction

Ensuring structural integrity is critical for public safety and long-term durability [1]. Early detection and timely repair of defects can prevent catastrophic failures and reduce maintenance costs [2]. The traditional defect inspection mainly relies on manual visual assessments, which are labor-intensive, time-consuming, and prone to human error [3]. With the advancement of deep learning, computer vision techniques have been widely adopted for structural defect detection, including classification [4], localization [5], and segmentation tasks [6]. Most existing approaches are grounded in supervised learning, which demonstrates high accuracy by learning complex defect patterns from labeled datasets [7]. However, the dependence on large-scale, high-quality labeled data poses a significant limitation for these existing approaches. The scarcity of such datasets significantly hinders model generalization and restricts real-world deployment [8].

In engineering practice, classification-level defect annotation (e.g., cracks, scaling, corrosion) is often sufficient for downstream applications. High-quality classification datasets offer two main benefits: (i) they serve as direct training resources for practical deployment scenarios [9]; and (ii) their availability for model pre-training facilitates transfer learning by improving feature representation and accelerates the weight updating of localization or segmentation detectors [10,11]. Nevertheless, the inefficient manual effort remains hindering the construction of such datasets. To alleviate the annotation burden, Zero-Shot Reasoning (ZSR) [12,13] and few-shot learning have emerged as promising alternatives [14]. ZSR leverages semantic priors, such as textual descriptions or class attributes, to recognize novel categories without labeled examples, while few-shot learning adapts to new tasks using only a limited number of annotated samples [15]. Although effective in general vision tasks, these strategies often suffer from poor generalization in structural defect detection due to domain mismatch and limited pre-training exposure.

Recent breakthroughs in Large Vision-Language Models (LVLMs), such as GPT-4 (OpenAI) [16] and Gemini (Google) [17], have demonstrated remarkable capabilities in multimodal understanding. Trained in large-scale image-text pairs, LVLMs can perform zero-shot classification, instruction-following, and chain-of-thought reasoning without task-specific fine-tuning [18]. These models exhibit agentic characteristics: they can follow natural language instructions, invoke external tools, and autonomously conduct complex visual reasoning, making them well-suited to act as intelligent annotation agents [19].

Building on this potential, an agentic framework, Agent-based Defect Pattern Tagger (ADPT), is proposed for automatic structural defect annotation in this study. ADPT integrates LVLMs with a pattern-matching semantic parser and an iterative self-questioning module that validates and refines predictions. The framework transforms unstructured LVLM outputs into standardized defect labels, enabling scalable, automated dataset construction with minimal human involvement. The main contributions are threefold: (i) propose a novel agentic framework for multi-class defect annotation without reliance on manually labeled data; (ii) design a self-questioning refinement mechanism that iteratively verifies and improves annotation accuracy; (iii) demonstrate the construction of high-quality defect classification datasets under both class-balanced and class-imbalanced conditions, supporting downstream tasks such as transfer learning and dataset bootstrapping.

The rest of this paper is structured as follows. Section 2 describes the overall architecture of ADPT, including LVLM prompting strategies and the annotation approach. Section 3 describes experimental setup and evaluation metrics. Section 4 presents performance analysis and comparative studies. Finally, Section 5 concludes the paper and discusses future research directions.

## 2 Methodology

### 2.1 Framework overview

The proposed agentic framework, ADPT focuses on filtering and extracting high-quality datasets from raw structural images encountered in real-world engineering practice. As shown in Fig. 1, the four core steps of ADPT are as follows:

Encoding phase: Raw images are resized to a fixed resolution using interpolation and normalized to standardize brightness and contrast, enhancing salient structural features. These preprocessed images are then encoded into Base64 strings and embedded into a lightweight JSON schema, ensuring compatibility with LVLM input requirements and minimizing transmission overhead. This representation serves as a unified interface for multimodal semantic processing.

Generation phase: Carefully designed textual prompts are combined with the encoded visual payload and fed into an LVLM. Leveraging its zero-shot reasoning capability, the model generates discriminative, human-readable descriptions of each image (e.g., "The image shows a textured surface that appears to be a wall with rectangular sections. The surface has patches of light blue color, along with areas of faded or peeling paint, indicating wear or damage…"). This process transforms raw visual inputs into structured semantic outputs without any manual labeling.

Annotating phase: A pattern-matching module parses the LVLM-generated descriptions and aligns relevant terms with predefined defect categories. Images are automatically classified and assigned to corresponding storage directories, enabling end-to-end defect annotation and category-level dataset construction.

Refinement phase: To validate and enhance annotation reliability, ADPT employs an iterative self-verification protocol. Each image-description pair is re-evaluated by the LVLM to assess consistency between the visual content and the generated text. Samples flagged as inconsistent are reassigned to an "uncertain" class for recursive refinement.

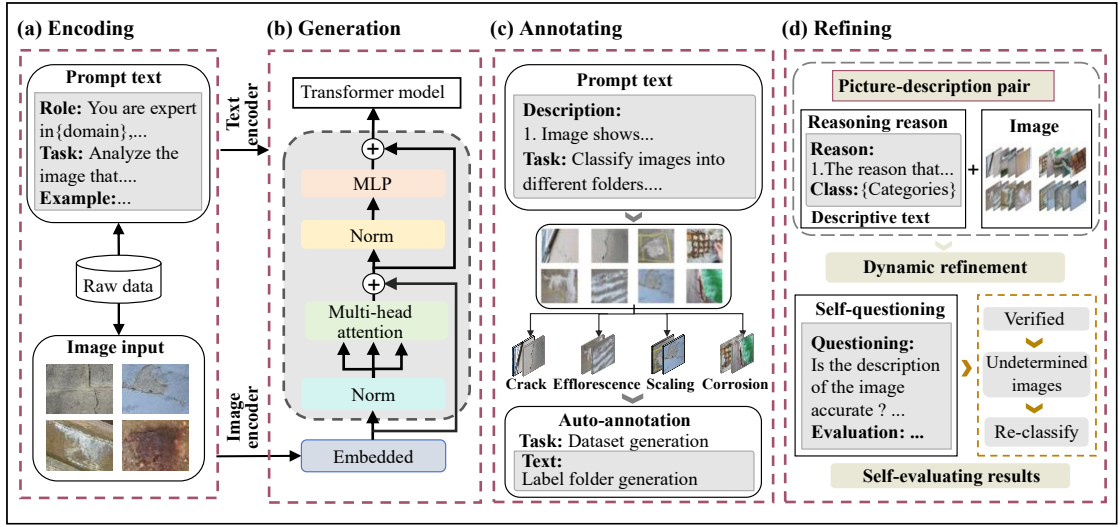

Fig. 1 Overall workflow of the proposed ADPT framework for fully automated structural defect detection and category-level annotation using LVLMs.

### 2.2 Different strategies for automatic annotation

Given the importance of prompt formulation, we implement two complementary annotation strategies within ADPT: a Zero-Shot Reasoning approach and a Feature-Prompt-Based (FPB) strategy. Both can be enhanced by the proposed Self-Questioning Refinement (SQR) module to further improve accuracy.

### 2.2.1 Zero-shot reasoning-based strategy

The ZSR strategy exploits the LVLM's ability to associate visual features with semantic categories without prior training. It maps each input image $I \in \mathbb{R}^{H \times W \times 3}$ into a normalized Base64-encoded form $\hat{I}$, wrapped in a JSON schema suitable for LVLM input. The model processes the image via its visual encoder $f_{img}$, producing a latent feature vector $v_{img} \in \mathbb{R}^d$. Instead of generating generic captions, the LVLM is prompted to produce a concise, category-relevant description: $V = (w_1, w_2, ..., w_n)$. This textual output is further encoded using the language encoder $f_{text}$, yielding a semantic vector $S_{img} = f_{text}(V)$. Classification is performed by computing cosine similarities between $S_{img}$ and predefined semantic prototypes $\{S_{c_i}\}_{i=1}^{5}$ for the categories (e.g., crack, corrosion, efflorescence, scaling, uncertain): $\hat{c} = \arg\max \cos(S_{img}, S_{c_i})$, where $\cos(S_{img}, S_{c_i}) = \frac{S_{img} \cdot S_{c_i}}{\|S_{img}\| \cdot \|S_{c_i}\|}$. Once the predicted class $\hat{c}$ is obtained, the system automatically creates the corresponding directory and stores the image. A structured

annotation log records the filename, LVLM-generated description, and predicted label, ensuring transparency and traceability.

**2.2.2 Feature-prompt-based strategy**

Unlike the ZSR strategy, FPB incorporates domain knowledge via explicit category-specific prompts. After preprocessing, the image is passed through $f_{img}$ to obtain $v'_{img}$. A feature prompt $P_i$ (e.g., "*Corrosion: rusty, eroded, or oxidized surfaces, especially on metal parts*") is prepended to steer the LVLM's attention toward discriminative defect features. This context-aware fusion generates a focused description: $V'=(w'_1, w'_2, ..., w'_n)$, which is again embedded as $S'_{img} = f_{text}(V')$, follow by classification: $\hat{c'} = \arg\max \cos(S'_{img}, S_{c_i})$. The remainder of the approach, including directory allocation and logging, is identical to the ZSR strategy. By guiding attention toward salient attributes, FPB improves semantic alignment and reduces confusion among visually similar defects.

**2.2.3 Self-questioning refinement**

To further improve robustness and eliminate semantic ambiguity, we introduce a SQR module. This mechanism evaluates the alignment between the image, its generated description, and the predicted label. Each image $I''$ is encoded as $B(I'')$, generating a semantic vector $S''_{img}$ and final label $\hat{c''}$. These components are combined into a structured prompt:

$$\{Img: B(I''); Text: S''_{img}; Label: \hat{c''}\}$$

The LVLM then re-evaluates this input, producing:

(i) An evaluation flag $\hat{e} \in \{Correct, Incorrect, Uncertain\}$, identified via keyword matching (e.g., "reasonable," "incorrect," "suggested change to"), indicating the validity of the original classification;

(ii) A rationale $R$, justifying the correction or supporting the original classification. If both $\hat{e} = Incorrect$ and $R$ contain explicit correction suggestions (e.g., "suggest changing to 'crack'"), the system flags the image label as semantically biased and moves it from its original category to a new folder for subsequent review. Likewise, any sample deemed uncertain, either due to ambiguous feedback or conflicting instructions, is isolated to prevent low-confidence annotations from contaminating the generated dataset. Through this filtering, only high-confidence examples advance, while ambiguous cases are preserved for targeted re-evaluation or manual inspection, ensuring a transparent audit trail that links each image to its self-questioning outcome and explanatory reasoning.

Beyond refining single-path predictions, the self-questioning module seamlessly supports a multi-path ensemble mechanism that pools assessments from both the zero-shot and FPB strategies. For each image, labels $\hat{c}_1$ and $\hat{c}_2$ generated by the two inference routes are independently subjected to self-questioning, yielding evaluation flags $\hat{e}_1$ and $\hat{e}_2$. These flags are then aggregated using a confidence-weighted voting scheme:

$$\hat{c} = \underset{c_i \in C}{\mathrm{argmax}} \sum_{k=1}^{2} \alpha_k \cdot \mathbf{1}[\hat{e}_k = \textit{Correct} \wedge \hat{c}_k = c_i],$$

where $\alpha_k$ represents the relative trustworthiness of each path and $\mathbf{1}[\cdot]$ is an indicator function.

By integrating instruction-following behavior with multi-level semantic alignment, the SQR module introduces a dynamic, self-auditing layer into the annotation workflow. This refinement step not only corrects biased or uncertain predictions in real time but also ensures transparency by maintaining a verifiable trace of each decision. As a result, the proposed framework gains significant robustness in automatic annotation tasks, particularly in scenarios involving complex or ambiguous defect patterns.

## 3 Experiment design

### 3.1 Overall experiment workflow

To evaluate the proposed framework's ability to perform structural defect annotation without labeled data, an end-to-end experiment was conducted incorporating both ZSR and FPB strategies, each optionally enhanced by the SQR module. As illustrated in Fig. 2, each image was first assessed for defect presence. If deemed defective, it was further classified into one of four categories: crack, efflorescence, scaling, or corrosion. To assess the influence of prompt design, synonym lists were constructed for each defect type and corresponding ZSR and FPB prompt design. Both strategies were evaluated using four Large Vision-Language Models (LVLMs): Gemini-2.0-Flash [20], GPT-4o-mini [21], Grok-4 [22] and Qwen-2.5-VL-32B [23]. The outputs were then refined through the SQR module, which re-evaluated uncertain samples and corrected low-confidence classifications.

Annotation accuracy is reported on both class-balanced and class-imbalanced datasets, enabling an assessment of robustness across data distributions. All results are collected and visualized to compare how the prompt design, reasoning style and evaluator feedback influence each model's performance.

To systematically investigate prompt sensitivity, three types of prompt vocabularies with varying semantic specificity are developed:

General Language (GL): common non-technical synonyms

Expert Terminology (ET): canonical defect terms

Technical Jargon (TJ): domain-specific descriptors

Tab. 1 presents the vocabulary sets used for the four target defect types. To ensure fair comparison, each input prompt only included one term from the selected vocabulary type, avoiding additional synonyms or contextual bias. Classification results were then compared across prompt sets to quantify the effect of semantic prior on LVLM performance.

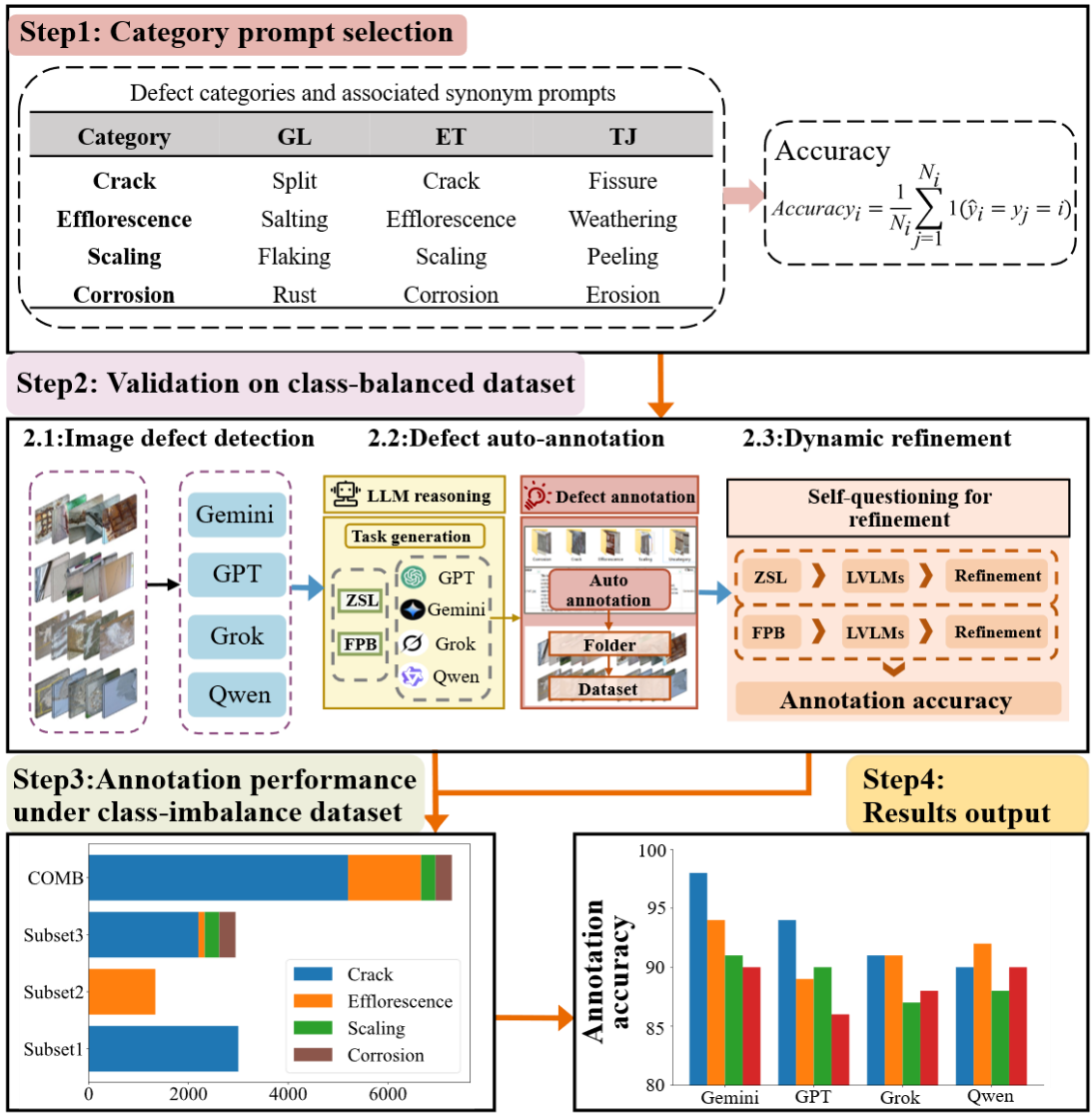

Fig. 2 Experimental workflow for automatic defect detection and multi-class annotation using the proposed framework.

Tab. 1 Defect categories and associated synonym prompts.

| Category | General Language (GL) | Expert Terminology (ET) | Technical Jargon (TJ) |
|---|---|---|---|
| **Crack** | Split | Crack | Fissure |
| **Efflorescence** | Salting | Efflorescence | Weathering |
| **Scaling** | Flaking | Scaling | Peeling |
| **Corrosion** | Rust | Corrosion | Erosion |

## 3.2 Dataset

To comprehensively assess the proposed framework under both controlled and real-world conditions, experiments were conducted on two distinct data settings: a class-balanced dataset and a class-imbalanced dataset.

The class-balanced dataset comprises 800 images, evenly split between defective and non-defective categories. Defective samples are drawn from multiple open-source repositories [24-28] and uniformly distributed across four common defect types, including crack, scaling, efflorescence, and corrosion, with 100 images per class. All images were retained at their original resolutions to preserve variability in visual characteristics such as texture, scale, and lighting. Representative samples are presented in Fig. 3.

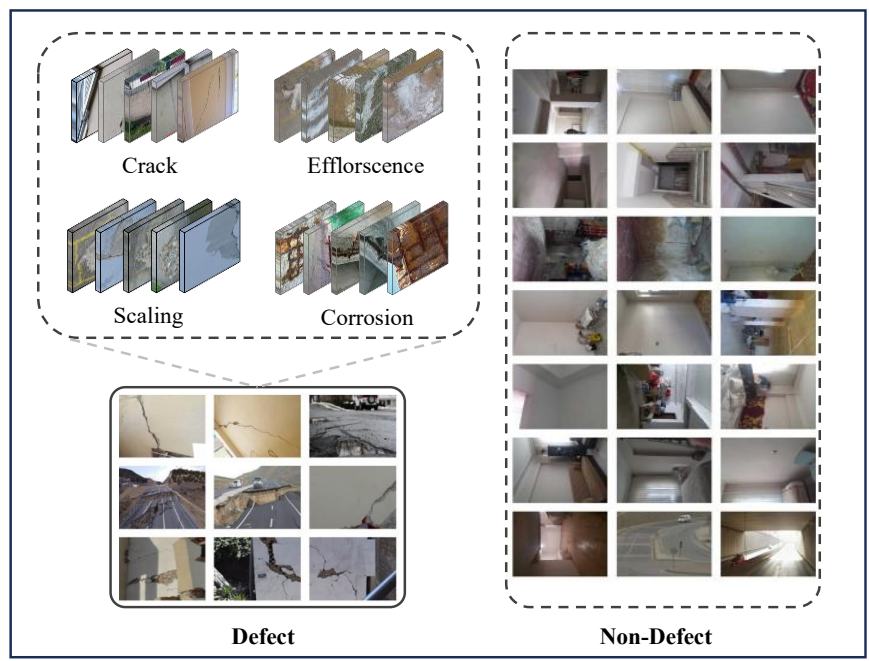

Fig. 3 Representative defect and non-defect images in the class-balanced dataset.

To simulate field-like conditions where defect distributions are naturally skewed, a class-imbalanced dataset was constructed by aggregating three heterogeneous sources. These included single-defect collections (e.g., 3000 crack images [29]; 967 efflorescence images [26]) and a multi-defect dataset containing variable quantities of all four defect types [25,27,28,30], shown in Fig. 4. This composite setting allowed for evaluating model robustness across both intra-class density variation and inter-class imbalance, which are typical challenges in practical infrastructure inspection scenarios.

In both datasets, all images are randomly shuffled and placed into a unified input directory. The framework sequentially processes each image using an agentic LVLM, which predicts the defect category and automatically routes the image to its corresponding output folder. Annotation performance is then quantified by comparing these predicted labels with ground truth, yielding accuracy metrics at both the per-class and overall levels.

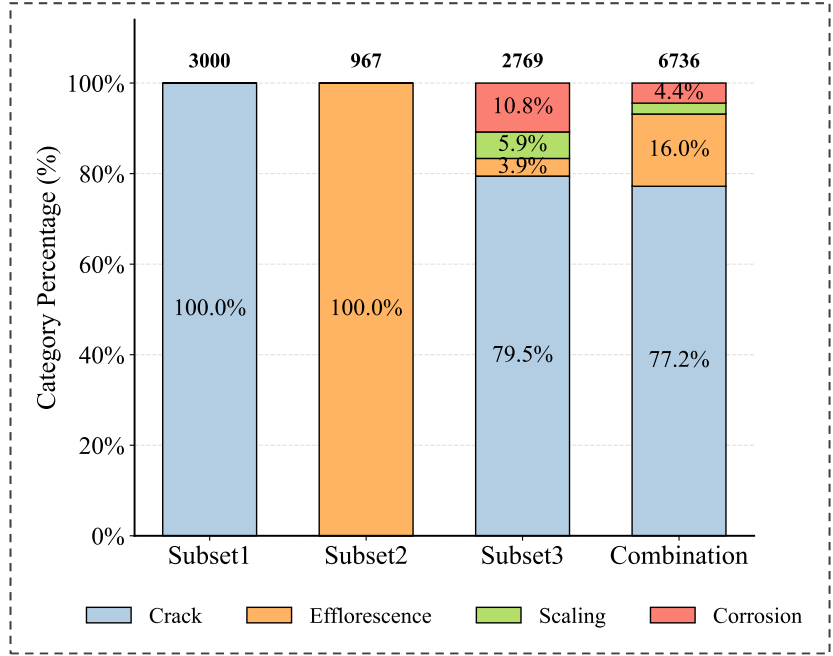

Fig. 4 Image distribution across class-imbalanced subsets and the composite dataset.

### 3.3 Evaluation metrics

While the framework may omit a few correctly classified samples due to uncertainty filtering, such omissions are acceptable in practical dataset construction, where precision outweighs recall. Hence, the classification accuracy is adopted as the primary evaluation metric, defined as:

$$Accuracy = \frac{Number\ of\ Correct\ Predictions}{Total\ Number\ of\ Predictions} \times 100\% \ .$$

To avoid information leakage, where the LVLM might infer labels from file names rather than image content, a uniform naming convention for all input images is employed. Each file name must include an underscore followed by a single-character code (e.g., "_A," "_B"), representing the true class label. During evaluation, the system extracts this character for validation, ensuring label integrity without exposing descriptive metadata.

A built-in parser verifies every prediction against the encoded ground truth, ensuring consistency across the evaluation set. This strict protocol guarantees that reported accuracy reflects genuine visual-semantic reasoning by the model, rather than reliance on spurious filename cues.

## 4 Results and discussion

### 4.1 Impact of defect category prompt design

The influence of prompt semantics on defect classification performance is examined by employing three types of category descriptors: GL, ET, and TJ. Each prompt type is applied to 100 representative images from four defect categories. The classification outcomes are illustrated in Fig. 5.

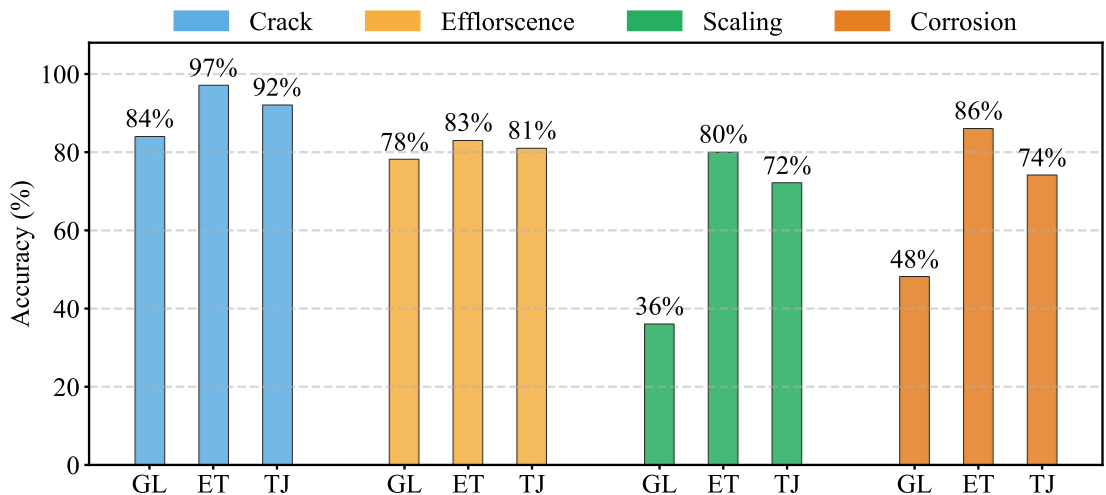

Fig. 5 Classification accuracy under the prompt designs of General Language, Expert Terminology, and Technical Jargon, for crack, efflorescence, scaling, and corrosion categories.

Accuracy obtained using ET prompts consistently outperforms that from TJ and GL across all categories. For example, crack identification reaches 97% under ET prompts, compared to 92% under TJ and 84% under GL. A similar pattern is observed for efflorescence and corrosion. In particular, classification performance for scaling is notably affected by prompt type, where only 36% accuracy is achieved with GL, while ET and TJ result in 80% and 72%, respectively.

These results demonstrate that domain-specific prompts improve semantic-visual alignment within LVLMs. In contrast, prompts based on everyday language tend to lack morphological specificity, leading to misclassification. The use of expert-curated terminology is thus considered essential for fine-grained defect annotation.

**4.2 Performance on class-balanced datasets**

**4.2.1 Binary defect classification**

Binary classification is performed on a balanced dataset comprising 400 defective and 400 non-defective images. The results are presented in confusion matrices shown in Fig. 6.

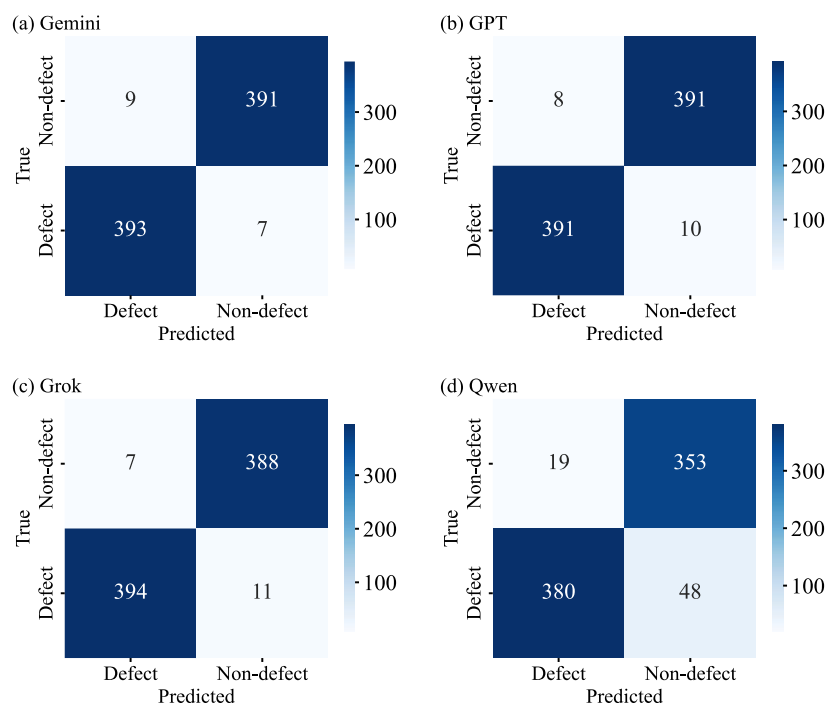

Fig. 6 Confusion matrices illustrating binary classification results of Gemini, GPT, Grok, and Qwen, for defective and non-defective images.

Among the evaluated models, Gemini exhibits the highest classification accuracy, with 393 defective and 391 non-defective images correctly identified. GPT and Grok achieve comparable performance, while Qwen records a lower accuracy due to a higher rate of false predictions. The disparity in false positive and false negative distributions across models suggests differing sensitivities and risk preferences, which may inform deployment decisions in specific operational contexts.

### 4.2.2 Multi-class defect annotation

To assess the capability of LVLMs in assigning defect-specific labels, classification tasks are extended to four categories: crack, efflorescence, scaling, and corrosion. Results under both ZSR and FPB strategies are presented in Fig. 7.

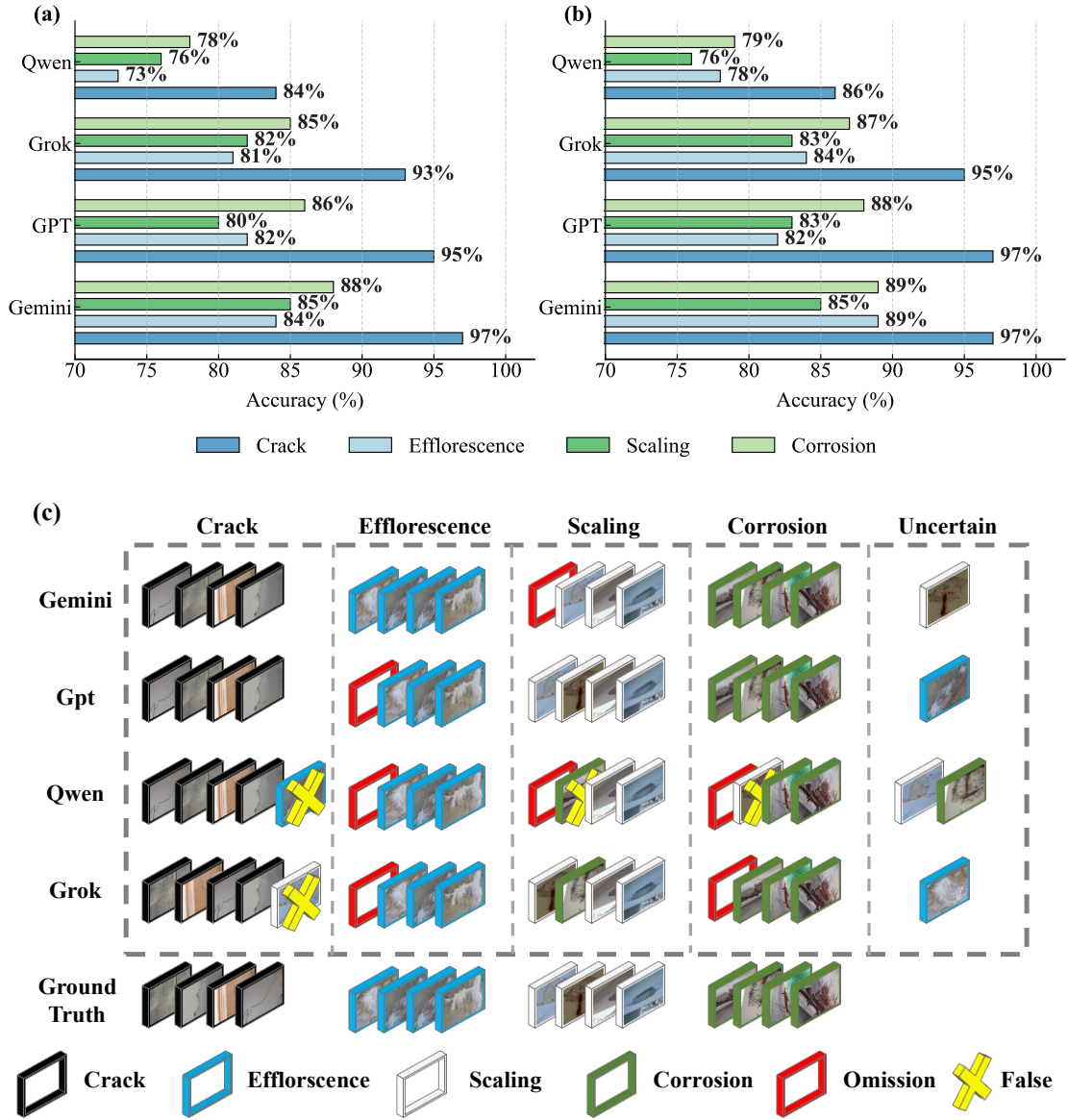

Fig. 7 Labeling accuracy of Gemini, GPT, Qwen, and Grok, for various defect types with the ZSR and FPB strategies: (a) ZSR strategy; (b) FPB strategy; (c) Representative examples of defect annotations generated using FPB-guided LVLMs.

Across all models, FPB consistently improves classification accuracy. The integration of task-oriented prompts, such as "layered material scaling," provides additional structural context, enhancing disambiguation between visually similar defects. Although classification of cracks reaches the highest accuracy (up to 97%), lower performance is observed for scaling and efflorescence due to their ambiguous visual cues and inconsistent lexical representations. Instances of misclassification primarily occur between scaling and corrosion, or between efflorescence and cracks, indicating overlapping feature spaces. These findings confirm the critical role of morphological precision in prompt design. Incorporation of descriptive geometric and material-specific cues is necessary to reduce ambiguity and guide LVLMs toward correct label assignment.

### 4.2.3 Annotation refinement via self-questioning

The SQR module is integrated into both ZSR and FPB approaches to evaluate its corrective potential. As shown in Fig. 8, the inclusion of SQR yields consistent accuracy improvements across all models and defect categories.

Accuracy gains are particularly notable in categories characterized by high visual ambiguity. The refinement process involves adversarial prompting and semantic reclassification, allowing uncertain predictions to undergo a secondary reasoning loop. This loop decouples superficial feature associations and enforces semantic verification, thereby reducing volatility and enhancing robustness in complex classification scenarios.

The effectiveness of the SQR module is attributed to its dual-function design: adversarial negation challenges are used to probe semantic assumptions, while low-confidence samples are re-evaluated using targeted contextual prompts. This structure operates post-inference uncertainty management and introduces interpretability into the LVLM-driven annotation process.

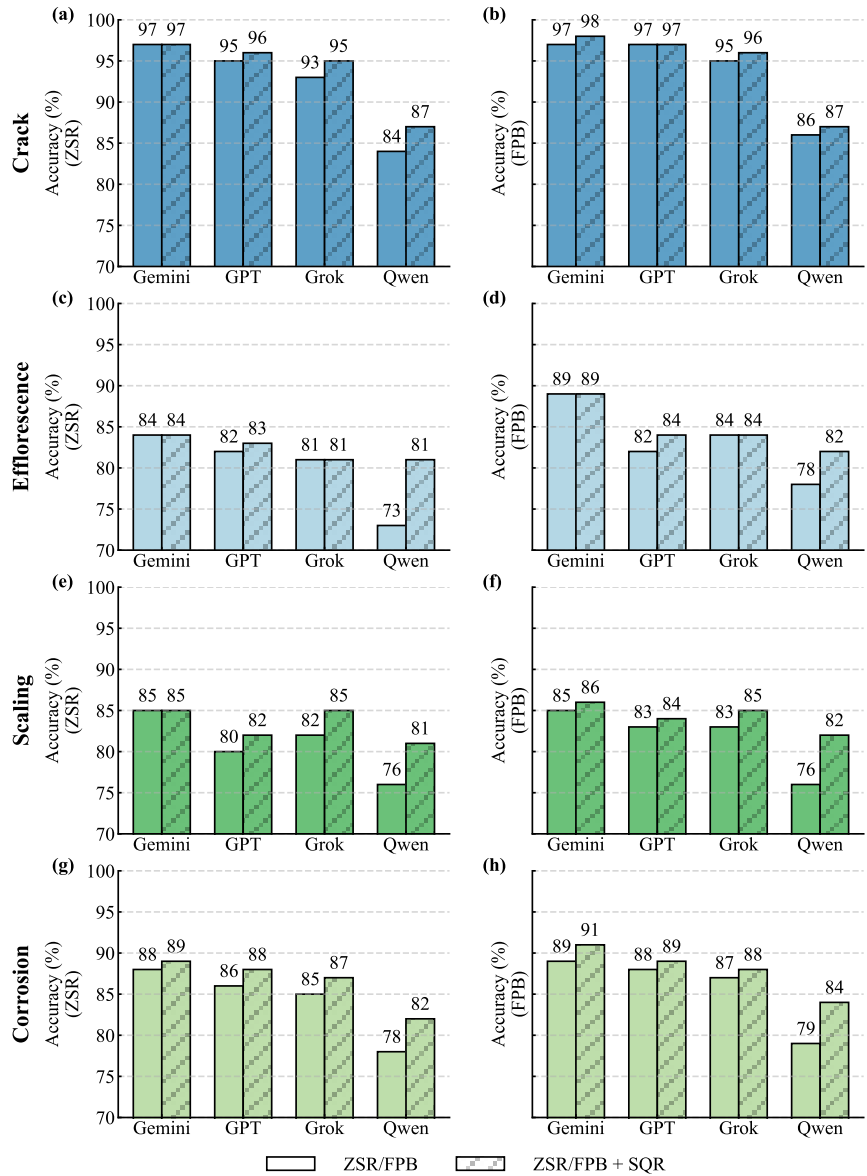

Fig. 8 Improvement in classification accuracy achieved by incorporating the Self-Questioning Refinement module.

### 4.3 Validation on class-imbalanced data

To further assess robustness, the framework is evaluated under class-imbalanced conditions reflective of field inspection data. Three heterogeneous sources, including the crack-only (Subset1), efflorescence-only (Subset2), and multi-defect-mixed (Subset3) collections, with their combination collection, are chosen for comparative study. The optimal configuration (Gemini + FPB + SQR) is applied, and results are summarized in Fig. 9.

Despite varying sample distributions and heterogeneity in defect features, the framework consistently maintains high accuracy. For single-class subsets, crack classification achieves 90.9% accuracy. Efflorescence, though more visually diffuse, is classified with 80.0% accuracy. When evaluated on the composite datasets containing all four defect types, per-class accuracies remain stable, with minimal performance degradation.

These findings further affirm that the framework generalizes effectively across imbalanced datasets. The classification process, driven by semantic reasoning rather than frequency bias, preserves consistency even when defect categories are underrepresented. The use of LVLMs enables multiscale feature extraction and robust category independence, contributing to scalability and real-world applications.

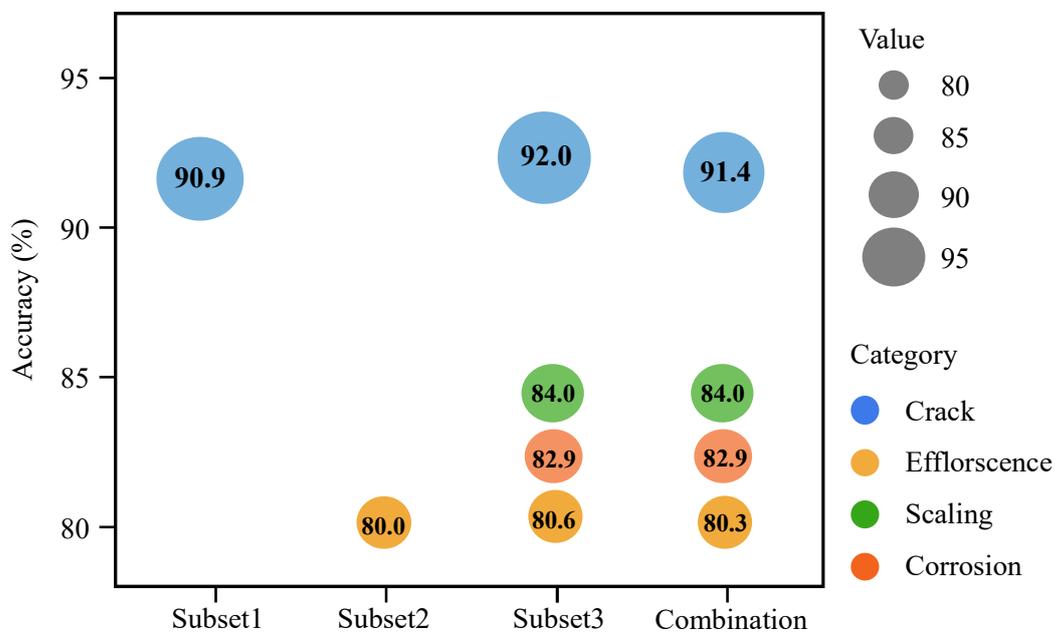

Fig. 9 Classification accuracy on crack-only (Subset1), efflorescence-only (Subset2), multi-defect-mixed (Subset3), and all-combined (Combination) class-imbalanced datasets for four defect categories.

## 5 Conclusions

The reliance on manual labeling in traditional structural defect inspection imposes considerable constraints on the deployment of deep-learning-based solutions in engineering domains. To address this challenge, an agentic annotation framework, ADPT, has been introduced, which leverages the reasoning capabilities of LVLMs, domain-specific prompt strategies, and a self-questioning refinement module.

Through a series of controlled evaluations, the efficacy of prompt design, reasoning strategy, and refinement mechanisms has been systematically validated. The use of expert terminology within prompts consistently yields superior annotation performance, surpassing general or technical synonyms by margins of up to 44%. Under class-balanced conditions, defect and non-defect discrimination accuracy reaches 98%, while multi-class annotation accuracy varies between 84% and 97%, depending on the defect type. Feature-prompt-based reasoning demonstrates consistent improvement over zero-shot approaches, particularly in visually ambiguous categories, while integration of the refinement module further enhances labeling precision by resolving boundary cases and semantic drift.

When applied to class-imbalanced datasets reflecting real-world defect distributions, annotation accuracy remains stable across diverse categories, ranging from approximately 80% to 92%, demonstrating strong generalization and robustness. These results confirm the viability of ADPT in supporting efficient and accurate image annotation workflows for structural inspection.

Despite these strengths, limitations remain. The framework's performance is inherently influenced by the pretraining data and architectural biases of proprietary LVLMs. Future extensions are envisioned to include prompt-tuning modules for adaptive generalization, and the integration of lightweight segmentation heads to enable pixel-level localization.

ADPT presents a scalable and transferable solution to the longstanding annotation bottleneck in structural defect recognition. Its agentic design and multimodal adaptability establish a promising foundation for broader adoption in safety-critical engineering applications.

## CRediT authorship contribution statement

**Sheng Jiang**: Validation, Investigation, Writing - original draft, Funding acquisition, Project administration. **Yuanmin Ning**: Investigation, Methodology, Visualization. **Bingxi Huang**: Investigation, Writing - review & editing. **Peiyin Chen**: Investigation,

Methodology, Visualization. **Zhaohui Chen**: Conceptualization, Methodology, Visualization, Investigation, Writing - review & editing.

## Acknowledgement

The work is supported by the National Natural Science Foundation of China (Grant No. 52479098).

## Declaration of generative AI and AI-assisted technologies in the writing process

During the preparation of this work the author(s) used GPT-5 in order to polish the language. After using this tool/service, the author(s) reviewed and edited the content as needed and took full responsibility for the content of the publication.